\documentclass{article} 
\usepackage{iclr2023_conference,times}

\usepackage{amsmath,amsfonts,bm}

\def\eqref#1{equation~\ref{#1}}

\def\1{\bm{1}}

\DeclareMathAlphabet{\mathsfit}{\encodingdefault}{\sfdefault}{m}{sl}
\SetMathAlphabet{\mathsfit}{bold}{\encodingdefault}{\sfdefault}{bx}{n}

\newcommand{\softmax}{\mathrm{softmax}}

\DeclareMathOperator*{\argmax}{arg\,max}

\usepackage{times}
\usepackage{latexsym}
\usepackage{times}
\usepackage{latexsym}
\usepackage{makecell}
\usepackage{graphicx}
\usepackage{hyperref}
\usepackage{wrapfig}
\usepackage{xcolor}

\usepackage[T1]{fontenc}

\usepackage[utf8]{inputenc}
\usepackage{breqn}
\usepackage{microtype}

\title{S2vNTM: Semi-supervised vMF Neural Topic Modeling}

\begin{document}
\maketitle
\begin{abstract}
Language model based methods are powerful techniques for text classification. However, the models have several shortcomings. (1) It is difficult to integrate human knowledge such as keywords. (2) It needs a lot of resources to train the models. (3) It relied on large text data to pretrain. In this paper, we propose Semi-Supervised vMF Neural Topic Modeling (S2vNTM) to overcome these difficulties. S2vNTM takes a few seed keywords as input for topics. S2vNTM leverages the pattern of keywords to identify potential topics, as well as optimize the quality of topics' keywords sets. Across a variety of datasets, S2vNTM outperforms existing semi-supervised topic modeling methods in classification accuracy with limited keywords provided. S2vNTM is at least twice as fast as baselines.

\end{abstract}

\section{Introduction}


Language Model (LM) pre-training~\cite{https://doi.org/10.48550/arxiv.1706.03762, https://doi.org/10.48550/arxiv.1810.04805} has proven to be useful in learning universal language representations. Recent language models such as ~\cite{yang2019xlnet, https://doi.org/10.48550/arxiv.1905.05583, https://doi.org/10.48550/arxiv.2201.08702, ding-etal-2021-ernie} have achieved amazing results in text classification. Most of these methods need enough high-quality labels to train. To make LM based methods work well when limited labels are available, few shot learning methods such as~\cite{Bianchi2021PretrainingIA,Meng2020DiscriminativeTM,Meng2020TextCU,Mekala2020ContextualizedWS,Yu2021FineTuningPL,Wang2021XClassTC} have been proposed. However, these methods rely on large pre-trained texts and can be biased to apply to a different environment.  

Topic modeling methods generate topics based on the pattern of words. To be specific, unsupervised topic modeling methods ~\cite{blei2003latent,teh2006hierarchical,miao2018discovering,dieng2020topic} discover the abstract topics that occur in a collection of documents. Recently developed neural topic modeling achieves faster inference in integrating topic modeling methods with deep neural networks and uncovers semantic relationship ~\cite{zhao2020neural, pmlr-v108-wang20c}. Compared to unsupervised topic modeling methods, semi-supervised topic modeling methods~\cite{mao2012sshlda,jagarlamudi-etal-2012-incorporating, gallagher2018anchored} allow the model to match the provided patterns from users such as keywords. However, these methods do not have high topic classification accuracy.

After studying topic modeling methods in real world applications~\cite{choi2017application, cao2019qos,kim2013mining,zhao2020targeted, xukdstm, xu2023vontss}, we realize the scenario that cannot be solved by current methods. The scenario involves topic exploration: users have identified a subset of topic keywords. They want to capture topics based on these keywords, while explore additional topics. They value the quality of the resulting topics and want to identify new topics while refining the topics' keywords iteratively~\cite{kim2013mining, smith2018closing}. In addition, users want to use the topic they created on topic classification.

\begin{figure*}
\hspace*{1em}\includegraphics[scale = 0.4]{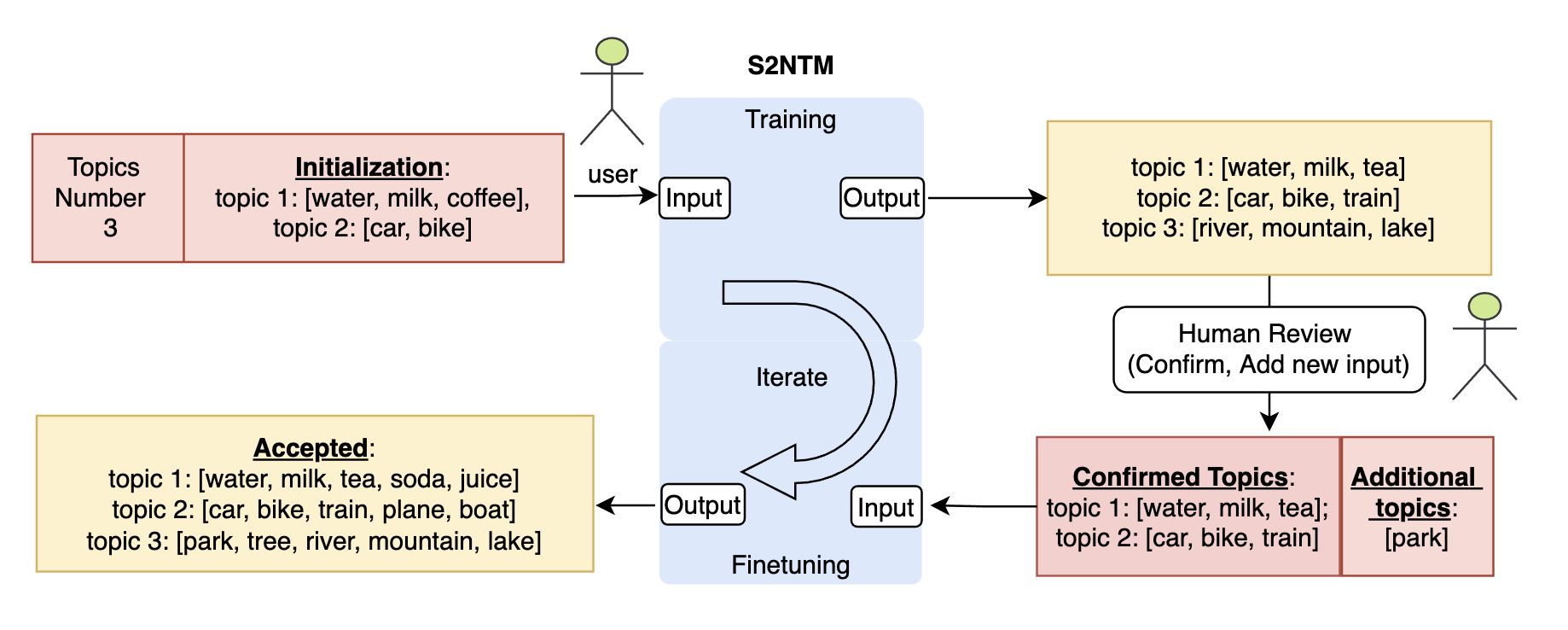}
\caption{\label{fig:people1}An S2vNTM application scenario. Human experts define topic keywords set and the number of topics first. During the training procedure, S2vNTM outputs keywords for each topic by merging the redundant keywords group and identifying new topics. Human experts then confirm/remove the keywords and/or add new keywords. S2vNTM continues refining the keyword list with a fast fine-tuning procedure. After a few iterations, S2vNTM provides users topics with high-quality keywords and high topic classification accuracy.}
\end{figure*}

In this work, we propose semi-supervised vMF neural topic modeling (S2vNTM). S2vNTM takes the desired number of topics as well as keywords/key phrases for some subsets of topics as input. It incorporates this information as guideline and leverages negative sampling to create topics that match the pattern of selected keywords. It creates additional topics which align with the semantic structure of the documents. It can help users remove redundant topics. Figure~\ref{fig:people1} illustrates how users interact with our model.
The advantages of this method include: \\
1. It consistently achieves the best topic classification performance on different datasets compared to similar methods. \\
2. S2vNTM only requires a few seed keywords per topic, and this makes it suitable for data scarce settings. It does not require any transfer learning.\\
3. S2vNTM is explainable and easy to fine-tune which makes it suitable for interfacing with subject-matter experts and low resource settings. \\

In sections below, we have shown Method in Section~\ref{Method} which describes the technical details of S2vNTM, Results in Section~\ref{results} and Conclusion and Future work in Section~\ref{conclusion}. Details on Modularity of S2vNTM is given in Appendix~\ref{Modularity}. Related Work and Challenges are described in Appendix~\ref{relatedwork}, Experiments in Appendix~\ref{experiments} and Ablation Studies in Appendix~\ref{ablation}.

\section{Method}
\label{Method}

\begin{figure*}
\hspace*{-1 em}
\includegraphics[scale = 0.5]{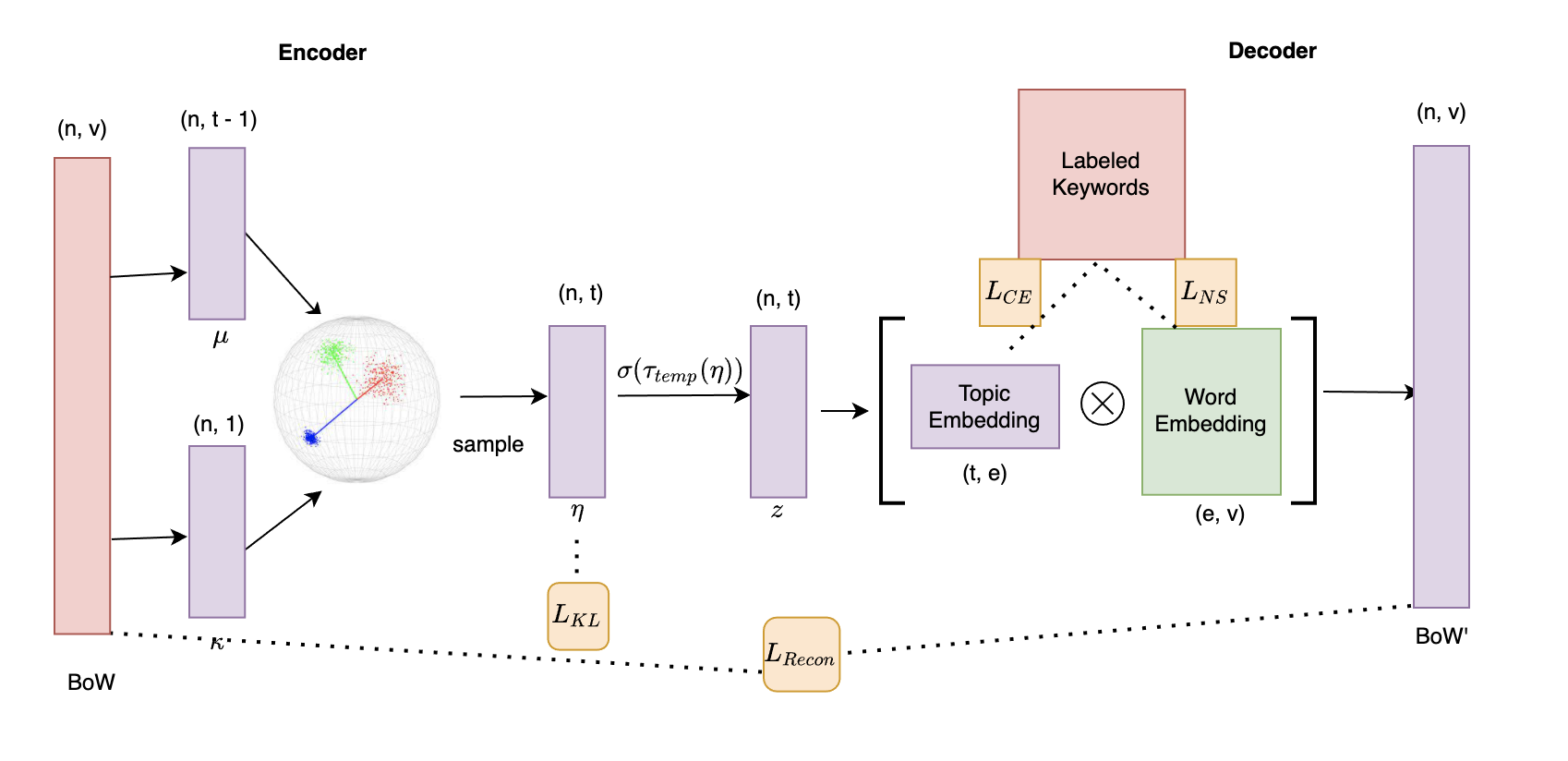} 
\caption{The neural network architecture of S2vNTM. We denote the dimension of the data in the bracket. $n$ is the number of documents. $v$ is the number of vocabularies. $t$ is the number of topics. $e$ is the dimension of embeddings. Word Embedding(green) is fixed during the training. Pink represents user provided data. Orange denotes all loss function including $L_{KL}$, $L_{Recon}$, $L_{CE}$ and $L_{NS}$}
\label{fig:people2}
\end{figure*}

Figure~\ref{fig:people2} shows the overall architecture of S2vNTM. The encoder is based on a Neural Topic Model leveraging von Mises-Fisher distribution. We use von Mises-Fisher distribution because it captures distributions on unit sphere and induces better clustering properties. To improve clustering, we add temperature function to the latent distribution(See details in Appendix~\ref{Temperature1}). The decoder tries to reconstruct the input from the topics while leveraging user-provided seeds for the topics. The model is trained end-to-end with the objective of minimizing reconstruction error while conforming to user-provided seeds and minimizing topic overlap.

\subsection{vNTM}
\label{vNTM}
We first introduce notation: the encoder network, $\phi$, encodes the bag of words representation of any document $X_{d}$ and outputs the parameters which can be used to sample the topic distribution $t_{d}$. The decoder is represented by a vocabulary embedding matrix $e_{W}$ and a topic embedding matrix $e_{t}$. We use a spherical word embedding ~\cite{meng2019spherical} trained on the dataset where we apply the model to create $e_{W}$ and keep it fixed during the training.  Spherical word embedding performs better on word similarity related tasks. If we do not keep embedding fixed, reconstruction loss will make the embeddings of co-occurred words closer which is not aligned with true word similarity. Fewer parameters to train can also make our method more stable. 
$W$ represents all selected vocabularies and $T$ contains all topics. In this notation, our algorithm can be described as follows: for every document $d$, (1) input bag of word representation $X_{d}$ to encoder $\phi$. (2) Using $\phi$, output direction parameter $\mu$ and variation parameter $\kappa$ for vMF distribution.\cite{xu2023vontss} (3) Based on $\mu$ and $\kappa$, generate a topic distribution $t_{d}$ using temperature function. (4) Reconstruct $X_d$ by $t_{d} \times \softmax(e_{t} e_{W}^{T})$. The goal of this model is to maximize the marginal likelihood of the documents: $\sum_{d = 1}^{D} \log p(X_{d} | e_{t}, e_{W})$. To make it tractable, the loss function combines reconstruction loss with KL divergence as below: \begin{equation} L_{Recon} = ( -E_{q_{\phi}(t_{d}|X_{d})}[log p_{\theta} (X_{d}|t_{d})] \end{equation}
\begin{equation}  L_{KL} = KL[q_{\phi}(t_{d}|X_{d}) || p(t_{d})] ) \end{equation} \textit{Our spherical word embedding is trained on the dataset without any pretraining. This can help embeddings deal with domain specific word. This can also make our model work for the language where there is not much text data available to pre-train. }
We leverage the vMF distribution as our latent distribution because of its clusterability and stability~\cite{xu2018spherical, 9547420, reisinger2010spherical, davidson2018hyperspherical}. Because of the design of the decoder, for each topic, it can be represented as a distribution of all words in vocabulary ($\softmax(e_{t} e_{W}^{T})$). \textit{When a document is provided, the user can identify the topics distribution of documents and also related keywords that contribute to these topics. Thus, the model is explainable.}

\subsection{Loss Function}
Our method allows users to define an arbitrary number of topics and provide keywords for some subsets of those topics. The model takes these two parameters as inputs and generates topics that include user's keywords as well as additional topics that align with topic distribution. With that being said, we 
want the prior loss similar to \begin{equation} L_{CE} = -\sum_{s \in S} max_{t \in T} \log \prod_{x \in s} q(x|t)\end{equation} where $S$ contains all keywords groups, s is a group of keywords and T is the group of topics, $q(x|t)$ stands for the probability of word x given t calculated by decoder. \begin{equation}q(x|t) = \frac{\exp{(e_{t_{j}} e_{x_{i}}^{T})}}{\sum_{x \in X} \exp{(e_{t_{j}} e_{x^{T}})}} \end{equation} This is the j-th row and i-th column of decoder embedding matrix $\softmax(e_{T} e_{W}^{T})$. Thus, it uses existed neural network structure to calculate and makes it computationally efficient.

\subsection{Topic and Keywords set Matching}
We want to make sure matched topics capture all documents related to the provided keywords. The problem of using $L_{CE}$ is that different keywords set may map to the same topic. It may merge the irrelevant topic set when that topic set is not aligned with most of the topics. To avoid this situation, we first select the topic that is most likely to align with this group of keywords but not align with words in all other groups. To be specific, we first select 
\begin{dmath} t_{s} = \argmax_{t\in T}(E_{x \in s} (\log q(x|t)) - \max_{x \in S } \log(q(x|t)) ) \end{dmath} This is inspired by Gumbel-Softmax \cite{jang2016categorical}. If one word in keywords set is dissimilar to the topic, the log will penalize it heavily and the topic is less likely to be matched. We also want to separate keyword groups which are different. If a keyword in another group has a higher probability in a topic, then $max_{s \in S} \log(q(s|t)) $ will be large, which makes the topic less likely to be the selected topic. If we have two similar keywords' sets, they can have similar and large $E_{x \in s} (\log q(x|t))$. These keywords sets can still map to the same topics. The benefit of this matching method is that it is more stable compare to method such as Gumbel-softmax and it can remove redundant topics by merging it with similar topics. 

\subsection{Negative Sampling}
We also want keywords as guidance to select other related keywords. Similar to ~\cite{yang2020mixed}, when a keyword set is matched with a topic, we want the topic to be less correlated with words that are unrelated to the matched keyword set. Thus, we leverage negative sampling. We first select the top N words in the selected topic using a decoder embedding matrix and sample each of top N word with sampling probability equal to $max_{x \in s} 1 - cos(x, x_{N})$ where $x_{N}$ stands for a word in top N words in that selected topic and $cos$ stands for cosine similarity. Our goal is to make words that are dissimilar to the provided keywords likely to be sampled, as seen in Table~\ref{tb_loss:table_of_figures1}. Negative Sampling can also help the model converge faster since it pushes away unrelated words quicker ~\cite{mimno-thompson-2017-strange}. The penalty we add for each keywords' set is:
\begin{equation} L_{NS, s} = \gamma \sum_{x \in ns} (\log (q(x | t_{s}))) \end{equation} where $ns$ contains words sampled from negative sampling. The loss of negative sampling is \begin{equation} L_{NS} = \sum_{s \in S} L_{NS, s} \end{equation} 
$\beta$ controls input keywords strength on overall loss function and $\gamma$ controls the strength of negative sampling. The overall loss function is: \begin{equation}L = L_{Recon} + L_{KL} + \beta *  L_{CE} + \gamma  * L_{NS} \end{equation} where $L_{NS}$ is the sum of all keywords set. $L_{Recon}$ is the reconstruction loss and $L_{KL}$ is the KL divergence loss. The benefit of this negative sampling design is that $q(x | t_{s})$ can be directly mapped from the decoder. Thus, it does not require additional computation, which saves computation resources.

\begin{table*}[h!]
\centering
\resizebox{\textwidth}{!}{%
\begin{tabular}{|l|lll|lll|lll|}
\hline
Model & AG News &    &  & R8 &    &  & DBLP &    &  \\ \hline
Metric & Accuracy  & Aucroc & Macro F1 & Accuracy  & Aucroc & Macro F1 & Accuracy  & Aucroc & Macro F1 \\ \hline
GuidedLDA & $0.734 \pm 0.037$  & $0.857 \pm 0.016$ & $0.735 \pm 0.039$ & $0.54 \pm 0.012$  & \textbf{$0.872\pm0.012$} & $0.309\pm 0.017$ & $0.493\pm0.009$  & $0.693\pm0.005$ & $0.47\pm0.008$ \\
CoreEx & $0.778\pm0.003$  & $0.889\pm0.001$ & $0.765\pm0.002$ & $0.532\pm 0.051$  & $0.762\pm0.025$ & \textbf{$0.394\pm0.024$} & $0.53\pm 0.009$  & \textbf{$0.8\pm0.005$} & $0.492\pm0.01$ \\
S2vNTM & \textbf{$0.795\pm0.009$}  & \textbf{$0.902\pm0.007$} & \textbf{$0.792\pm0.009$} & \textbf{$0.651\pm0.03$} &  $0.813\pm0.022$ & $0.362\pm0.049$ & \textbf{$0.598\pm0.029$}  & $0.793\pm0.022$ & \textbf{$0.545\pm0.032$} \\ \hline
\end{tabular}%
}

\caption{Scores and Standard Deviation for Accuracy, Macro F1 and Aucroc of GuidedLDA, CoreEx and S2vNTM models on AG News, R8 and DBLP datasets.}
\label{MainMetricsTable}
\end{table*}



\section{Results}
\label{results}
We ran our experiments 10 times with different seeds and show the result in Table~\ref{MainMetricsTable} (and Figure~\ref{fig:main_metrics} in the Appendix). (1) S2vNTM achieves the best accuracy in all three datasets. In fact, the worst reported accuracy of S2vNTM is higher than the best from the other two methods. We believe there are 3 reasons contributing to its superior performance. (i) It has high clusterability using vMF as a latent distribution. This makes our method easily clustered. (ii) Negative sampling excludes unrelated keywords from the topics. This makes our method perform better on documents that are related to keywords. (iii) S2vNTM also uses word embedding trained on the dataset. This makes our method perform well on documents that have words that are similar to words in keywords set. (2) S2vNTM keywords make more sense qualitatively in Table~\ref{tb_loss:table_of_figures1} in Appendix. This is due to KL divergence loss. Flexible concentration parameter $\kappa$ makes our method more locally concentrated. This makes topics different from each other. (3) S2vNTM also has a higher aucroc and Macro F1 score than other methods in most cases (from Table~\ref{MainMetricsTable}). This means that our method can deal with imbalanced datasets and can easily distinguish between classes. However, it performs less well on R8, which has 8 imbalanced classes. For class with less than 300 documents, keywords selected by tf-idf are less representative. Thus, it has lower performance and higher variance. Besides, our method using vMF distribution which has higher reconstruction loss when the dimension is high. R8 has 8 classes which make our method perform worse.

Qualitatively, as you can see in Table~\ref{tb_loss:table_of_figures1} , negative sampling reduces the importance of unrelated keywords such as \textit{call, york, company} while increasing the importance of given keywords such as \textit{military, industry, athlete}. Also, semantically, keywords in each set are closer to each other. For example, in the first set of keywords, \textit{government, war} are semantically more related to \textit{crime, rule} compared to \textit{call, election}. On the other hand, even if CorEx has good topic diversity, the keywords set is not coherent. For example, the last group in Table~\ref{qualitative} has \textit{inc, corp, people, bush, million} in one group. Determining the relationship between these keywords is not obvious.

\textbf{Speed} 
We run each model 10 times on AG News with different seeds to evaluate how long it takes to fine-tune the model by modifying 20 percent of keywords set. The average fine-tune time for our method is 51.33 seconds. To compare,  CatE~\cite{meng2018weaklysupervised} takes 888.61 seconds to fine-tune, while CorEx takes 94.98 seconds to fine-tune. This shows that our method is better suitable for iterative topic learning \cite{hu2014interactive} and resource restrictive environments.

Overall, qualitative results show that \textit{S2vNTM can help users find more coherent and relevant keywords compare to existed methods. Negative sampling makes the topics set more coherent. S2vNTM is at least twice faster than baselines.}

\begin{table*}[h!]
\scalebox{0.9}{

\begin{tabular}{ |c|c| } 
\hline
S2vNTM & S2vNTM + Negative Sampling \\
\hline
\textbf{government}, \textbf{war},  president, call, election &
\textbf{government}, \textbf{war},  \textbf{military}, crime, rule 
\\
\hline
\textbf{stock}, high, investor, \textbf{market}, york &
\textbf{stock}, investor, \textbf{market}, share, \textbf{industry} 
\\
\hline
\textbf{software}, \textbf{computer}, system, microsoft, company &
\textbf{software}, \textbf{computer}, microsoft,  system, technology  \\
\hline
game, sport, champion, season, team & 
game, sport, champion, season, \textbf{athelete}\\
\hline
united, reuters, international, state, union &
reuters, united, state, international, plan \\
\hline
reuters, report, target, http, company & 
reuters, report, target, http, company \\
\hline
\end{tabular} }

\caption{Comparison of top 5 keywords from each topics on AG News. The keywords that are given are [government,military,war], [stock,market,industry], [computer,telescope,software], [basketball,football,athlete].}
\label{tb_loss:table_of_figures1}
\end{table*}

\section{Conclusion and Future Work}
\label{conclusion}
In conclusion, we propose S2vNTM as an approach to integrate keywords as pattern to current neural topic modeling methods. It is based on vMF distribution, negative sampling, modified topic keywords mapping and spherical word embeddings. 
Our method achieves better classification performance compared to existing semi-supervised topic modeling methods. It is not sensitive to parameters. S2vNTM gives more coherent topics qualitatively. It also performs well when the input keywords set is less common in the dataset. It is also fast to fine-tune. It does not require pretraining or transfer learning. It only needs a few sets of seed words as input. 

The ablation study shows the potential of our method to further improve. In the future, we will focus on decreasing the gap between loss function and classification metric, incorporating sequential information and further improving the stability of the model. We will also work on improving its expressability in higher dimensions.

\newpage

\cleardoublepage

\bibliography{custom}
\bibliographystyle{acl_natbib}
\newpage
\appendix

\textbf{\large Appendix}
\label{appendix}

\section{Modularity of S2vNTM}
\label{Modularity}
Our methods can be plugged into variational autoencoder based topic modeling methods such as NVDM \cite{miao2016neural} and NSTM \cite{zhao2020neural}.
For NVDM, since their decoder is a multinomial logistic regression, we can consider that as the distribution of word over the topic. For $L_{CE}$, we can change $P(e_{w}|e_{t})$ to $P_{\theta} (x_{i}|h)$ (formula (6) in \cite{miao2016neural}) as it also represents the probability of certain word given all other words. For $L_{NS}$, we just sample it the same way as \cite{mikolov2013distributed}. For NSTM, since they also maintain topics and word embeddings (They name it G and E in the paper), we can use cosine similarity of these embeddings to create the loss functions $L_{CE}$ and $L_{NS}$ respectively. For that being said, this work can be easily extended by existed unsupervised neural topic modeling methods. 

\subsection{Temperature function}
\label{Temperature1}
Step (3) in Section~\ref{vNTM} introduced the concept of a temperature function. Temperature is a function that applies to  the sample generated by vMF distribution to form a topic distribution. To be specific, \begin{equation} t_{d} = \softmax(\tau_{temp}(\eta_{d})) \end{equation} where $\eta_{d}$ is the vector of sampled vMF distribution.  Since the sample from vMF is on the surface of a sphere, we have \begin{equation} \sum (\eta_{d}^{2}) = 1 \end{equation} In cases where the number of topics equals to 10, the most polarized $\eta_{d}$ is $(1, 0, 0, 0, 0, ...)$. If we apply softmax to this $\eta_{d}$, the highest topic proportion is 0.23, making latent space entangled and limit the clusterability. 

To overcome the expressibility concern mentioned in Related Work in Appendix \ref{vMF:dist:sec}, temperature function $\tau_{temp}$ is used to increase expressibility. For example, if we let $\tau_{temp}(\eta_{d}) = 10 * \eta_{d}$, the highest topic proportion of the above example becomes 0.99. This makes the produced topics more clustered. Also, we make  $\kappa$ flexible. The KL divergence of vMF distribution makes the distribution more concentrated while not influence the direction of latent distribution.

\section{Related Work and Challenges}
\label{relatedwork}

In this section, we touch on key concepts utilized in S2vNTM and their limitations. 

\subsection{Weakly-supervised text classification} 
Weakly supervised text classification methods aim to predict labels of texts using limited or noisy labels. Given class names, \cite{Wang2021XClassTC}  first estimates class representations by adding the most similar word to each class. It then obtains document representation by averaging contextualized word representations. Finally, it picks the most confident cdocuments from each cluster to train a text classifier. \cite{Yu2021FineTuningPL} improves weakly text classification on existed LM using contrastive regularization and confidence based reweighting. \cite{Meng2020TextCU} associates semantically related words with the label names. It then finds category-indicative words and trains the model to predict their implied categories. Finally, it generalizes the model via self-training. However, all these methods are time consuming to train and fine-tune which make it hard to be interactive. It is also hard to explain the reason behind certain classification. 

\subsection{Topic Modeling} Latent Dirichlet Allocation (LDA) \cite{blei2003latent} is the most fundamental topic modeling approach based on Bayesian inference on Markov chain Monte Carlo (MCMC) and variational inference; however, it is hard to be expressive or capture large vocabularies. It is time consuming to train the model.  It also has the tendency to identify obvious and superficial aspects of a corpus \cite{jagarlamudi-etal-2012-incorporating}  Neural topic model \cite{miao2018discovering}(NTM) leverages an autoencoder \cite{kingma2014semi} framework to approximate intractable distributions over latent variables which makes the training faster. To increase semantic relationship with topics, Embedded topic model (ETM) \cite{dieng2020topic} uses it during the decoder/reconstruction process to make topic more coherent and reduces the influence of stop words. However, the generated topics are not well clustered. Besides, using pre-trained embeddings cannot help the model identify domain specific topics. For example, topics related to Covid cannot be identified easily using pre-trained Glove embeddings~\cite{pennington-etal-2014-glove} since Covid is not in the embeddings.  To improve clusterability\cite{guu2018generating}, NSTM \cite{zhao2020neural} uses optimal transport to replace KL divergence to improve clusterability. It learns the topic distribution of a document by directly minimizing its optimal transport distance to the document’s word distributions. Importantly, the cost matrix of the optimal transport distance models the weights between topics and words, which is constructed by the distances between topics and words in an embedding space. Due to the instability of latent distribution, it makes it difficult to integrate external knowledge into these models. Existed semi-supervised NTM methods either are not stable \cite{wang2021neural,Harandizadeh_2022} or need specific twists~\cite{gemp2019weakly}.

\subsection{Semi-supervised Topic Modeling}
Semi-supervised Topic Modeling methods take few keyword sets as input and create topics based on these keyword sets. Correlation Explanation (CorEx) \cite{gallagher2018anchored} is an information theoretic approach to learn latent topics over documents. It searches for topics that are "maximally informative" about a set of documents. To be specific, the topic is defined as group of words and trained to minimize total correlation or multivariate mutual information of documents conditioned on topics. CorEx also accepts keywords by add a regularization term for maximizing total correlation between that group of keywords to a given topics. There is a trade-off between total correlation between documents conditioned on topics and total correlation between keywords to topics. GuidedLDA \cite{jagarlamudi-etal-2012-incorporating} incorporates keywords by combining two techniques. The first one defines topics as a mixture of a seed topic and a regular topic where topic distribution only generates words from a group of keywords. The second one associates each group of keywords with a Multinomial distribution over the regular topics. It transfers the keywords information from words into the documents that contain them by first sampling a seed set and then using its group-topic distribution as prior to draw the document-topic distribution. However, both methods fail to capture the semantic relationship between words. This means that when the provided keywords are less frequent in the corpus, the model's performance drop sharply.  

\subsection{Negative Sampling} Negative Sampling \cite{mikolov2013distributed} is proposed as a simplified version of noise contrastive estimation \cite{mnih2013learning}. It is an efficient way to compute the partition function of an non-normalized distribution to accelerate the training of word2vec. \cite{mikolov2013distributed} sets the negative sampling distribution proportional to the $\frac{3}{4}$ power of degree by tuning the parameters. Uncertainty based negative sampling \cite{li2013bootstrapping} selects the most informative negative pairs and iteratively updates how informative those pairs are. Some methods \cite{bucher2016hard} also account for the intra-class correlation. Negative sampling  is used in topic modeling methods since it can leverage the word-context semantic relationships \cite{shi2018short} or generate more diverse topics \cite{wu-etal-2020-short}. Both methods are applied in fully unsupervised scenario. In general, it needs to compute the similarity between the topic and all vocabularies. This step adds additional time and space complexity to the model which makes related methods less practical.

\subsection{von Mises-Fisher based methods} \label{vMF:dist:sec}

In low dimensions, the gaussian density presents a concentrated probability mass around the origin. This is problematic when the data is partitioned into multiple clusters. An ideal prior should be non informative and uniform over the parameter space. Thus, the von Mises-Fisher(vMF) is used in VAE. vMF is a distribution on the (M-1)-dimensional sphere in $R^{M}$, parameterized by $\mu \in R^{M}$ where $||\mu|| = 1$ and a concentration parameter $\kappa \in R_{\geq 0}$. The probability density function of the vMF distribution for $t \in R^{D}$ is defined as:
$$q(t|\mu, \kappa) = C_{M}(\kappa) \exp(\kappa\mu^{T}t)$$
$$C_{M}(\kappa) = \frac{\kappa^{\frac{M}{2} - 1}}{(2\pi)^{\frac{M}{2}} I_{\frac{M}{2} - 1}(\kappa)} + \log 2$$
where $I_{v}$ denotes the modified Bessel function of the first kind at order v. The KL divergence with vMF(., 0) \cite{davidson2018hyperspherical} is
$$KL(vMF(\mu, \kappa)|vMF(.,0)) = \kappa\frac{I_{\frac{M}{2}}(\kappa)}{I_{\frac{M}{2}-1}(\kappa)} $$
$$+ (\frac{M}{2} - 1) \log \kappa - \frac{M}{2} \log (2\pi)  - \log I_{\frac{M}{2}-1}(\kappa) $$ $$+ \frac{M}{2} \log \pi + \log 2 + \log \Gamma(\frac{M}{2})$$
vMF based VAE has better clusterability of data points especially in low dimensions \cite{guu2018generating}.

\cite{xu2018spherical} proposes using vMF(.,0) in place of Gaussian as $p(Z)$, avoiding entanglement in the center. They also approximate the posterior $q_{\phi}(Z|X)$ = $vMF(Z;\mu,\kappa)$  where $\kappa$ is fixed to avoid posterior collapse. The above approach does not work well for two reasons. First of all, fixing $\kappa$ causes KL divergence to be constant which reduces the regularization effect and increases the variance of latent distribution.  Another concern with vMF distribution is its limited expressability when its sample is translated into a probability vector. Due to the unit constraint, $softmax$ of any sample of vMF will not result in high probability on any topic even under strong direction $\mu$. For example, when topic dimension $M$ equals to 10, the highest topic proportion of a certain topic is 0.23.

\section{Experiments}
\label{experiments}
In this section, we report experimental results for S2vNTM and show that it performs significant better compared to two baselines.

\textbf{Datasets:}
We use three datasets: DBLP \cite{DBLP:conf/ijcai/PanWZZW16}, AG News \cite{zhang2016characterlevel}, R8 \cite{Lewis1997Reuters21578TC}. These datasets are all labeled.
AG News has 4 classes and 30000 documents per class with an average of 45 words per document. We select AG News since it is a standard dataset for semi-supervised topic modeling evaluation. DBLP has 4 classes. Documents per class varies from 4763 to 20890. Average document length is 5.4. We select DBLP to see how our model performs when document is short and categories are unbalanced. R8 is a subset of the Reuters 21578 dataset, which consists of 7674 documents from 8 different reviews groups. We select R8 dataset to see how our model performs when the number of keywords set and topics are large. We use the same keywords as \cite{meng2018weaklysupervised} for our experiments for AG News. For others, we use 20 percent of corpus as the training set to get our keywords by tf-idf score for each classes. To form the vocabulary, we keep all words that appear more than 15 times depending on the size of the dataset. We remove documents that are less than 2 words. We also remove stop words, digits, time and symbols from vocabulary. We also include bigram and trigram that appear more than 15 times.

\textbf{Settings:} The hyperparameter setting used for all baseline models and vNTM are similar to \cite{JMLR:v20:18-569}. We use a fully-connected neural network with two hidden layers of [256, 64] unit and ReLU as the activation function followed by a dropout layer (rate = 0.5). We use Adam~\cite{kingma2017adam} as optimizer with learning rate 0.002 and use batch size 256. We use \cite{smith2018superconvergence} as scheduler and use learning rate 0.01 for maximally iterations equal to 50. We use 50 dimension embeddings \cite{meng2019spherical} trained on the dataset where we apply the model. We set the number of topics equal to the number of classes plus one. Our code is written in pytorch and all the models are trained on AWS using ml.p2.8xlarge (NVIDIA K80). We use 80 percent data as test set.

\textbf{Baselines:} We compare our methods with GuidedLDA \cite{jagarlamudi-etal-2012-incorporating} and CorEx \cite{gallagher2018anchored}. CorEx are finetuned by anchor strength from 1 to 7 with step equal to 1 on the training set. GuidedLDA is finetuned using best seed confidence from 0 to 1 with step equal to 0.05 on the training set.

\textbf{Metrics:} To evaluate the classification performance of these models, we report \textbf{Accuracy}, \textbf{Macro F1} and \textbf{AUC}. We omit micro f1 since most of classes in these datasets are balanced and micro f1 is very similar to accuracy.

In addition, we want keywords in each topic to be diverse. This can help users to explore and identify new topics. We define \textbf{Topic Diversity} to be the percentage of unique words in the top 25 words of all topics~\cite{dieng2020topic}. Diversity close to 0 indicates redundant topics while diversity close to 1 indicates more varied topics.

\section{Qualitative Study}

\label{qualitative}
\scalebox{0.9}{

\begin{tabular}{ |c|c|c|c| } 
\hline
GuidedLDA & CorEx \\
\hline
iraq, kill, reuters, president, minister &
\textbf{government}, \textbf{war}, \textbf{military}, iraq, kill \\
\hline
reuters, \textbf{stock}, oil, price, profit &
\textbf{stock}, \textbf{market}, \textbf{industry}, price, oil \\
\hline
microsoft, company, \textbf{software}, service, internet & \textbf{software}, \textbf{computer}, microsoft, internet, service \\
\hline
win, game, team, season, lead & \textbf{footable}, \textbf{basketball}, game, win, season
\\
\hline
space, reuters, win, quot, world &
court, executive, chief, commission, union\\
\hline
quot, year, company, million, plan & 
inc, corp, people, bush, million\\
\hline
\end{tabular}
}

\section{Ablation Studies}
\label{ablation}
In this section we analyze and investigate the effect of various techniques and hyperparameters  on S2vNTM. We use AG News as the dataset since it is standard and has balanced classes. We run each experiment 10 times and report the barplot. Specifically, for parameters we analyze: 1. Number of topics 2. Different keyword sets 3. Temperature function 4. $\gamma$ ($L_{NS}$  multiplier) for topic modeling. For techniques, we analyze 1. Batch normalization and dropout and 2. Learnable distribution temperature. The first two are reported here and the rest are discussed in the Appendix.

\subsection{Effect of Number of topics}
\begin{figure*}
\hspace*{-8em}\includegraphics[scale=0.77]{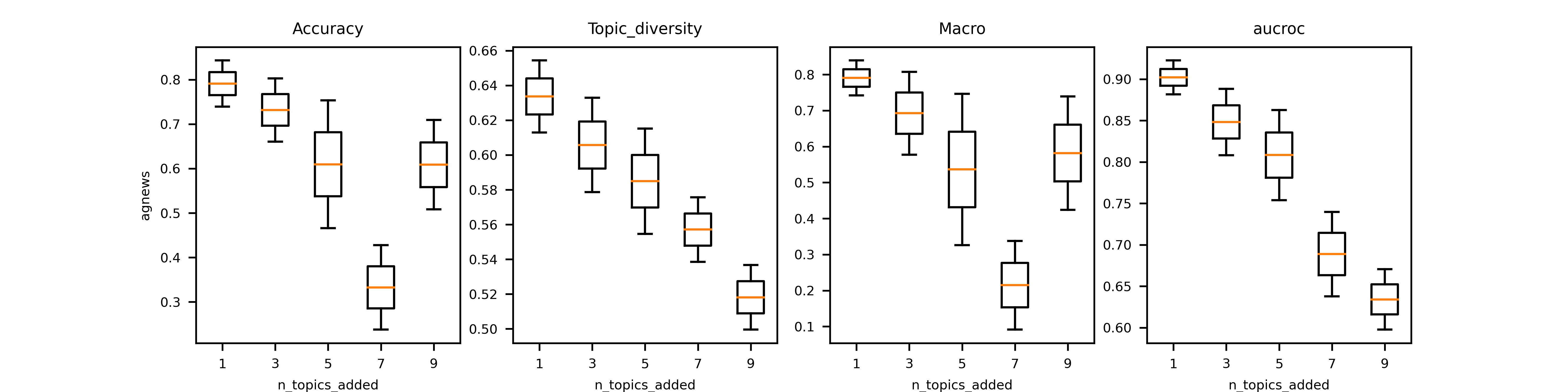}
\caption{Number of topics added for diversity VS various metrics for S2vNTM model. The topics is increasing from 5 to 13 which is increasing 1 to 9 from 4 basic topics}
\label{fig:ntopics_exp}
\end{figure*}

In this section, we analyze the effect of increasing in the number of topics from 5 to 13 shown in Figure~\ref{fig:ntopics_exp}. We see that the accuracy drops as the number of topics increases. This is because with increased number of topics, there is an increase in probability of adding additional topics that are similar to anchored topics and so the model gets confused while assigning words to topics. This could be either because of lack of topics in dataset or the latent space becoming very crowded i.e. space between vectors is less so it becomes difficult for models to discriminate between topics. Also, it seems that vMF based variational autoencoder performs less well in high dimension data. This could be addressed with an increase in distribution temperature discussed in Appendix~\ref{effect_distribution_temp}. 

\subsection{Effect of different keywords sets:} 
\begin{figure*}
\centering
\includegraphics[scale=0.5]{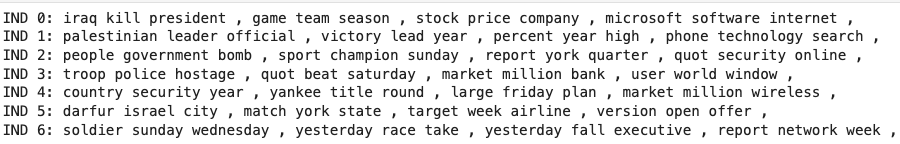}

\caption{New seed topic labels use. Figure~\ref{fig:newLabelsMetrics} reports classifcation metrics for these seeds.}
\label{fig:newLabels}
\end{figure*}

\begin{figure*}
\hspace*{-14em}\includegraphics[scale=0.55]{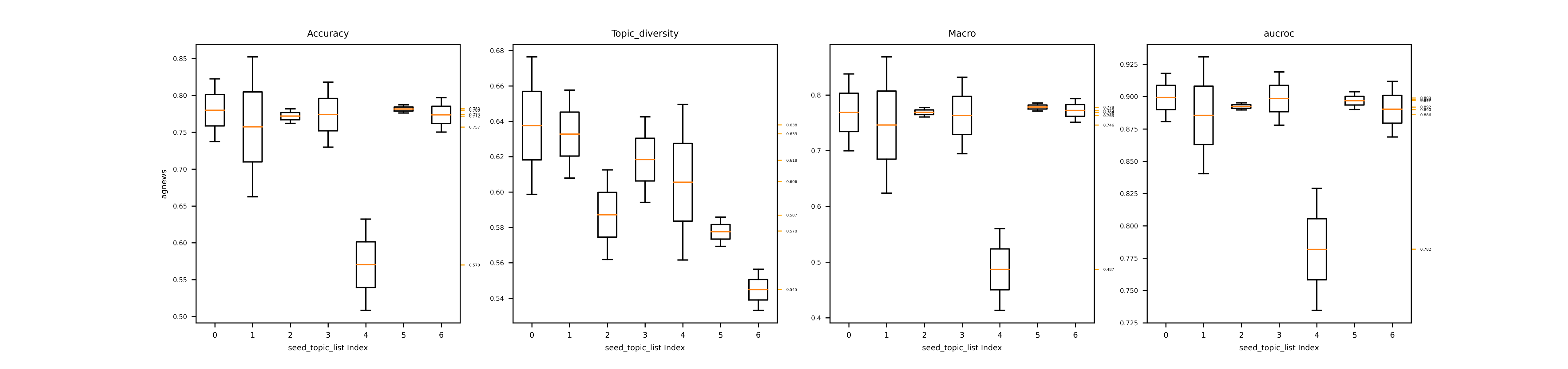}

\caption{Effect of different keywords sets listed in Figure~\ref{fig:newLabels} on classification and diversity metrics. Y-axis on right shows mean.}
\label{fig:newLabelsMetrics}
\end{figure*}

For traditional method such as CorEx and GuidedLDA, their performance drops when less frequent words are selected as keywords. To check the performance of our method on the less frequent keywords, we select top 30 keywords based on tf-idf score. Then we sort them based on frequency. The keywords set is shown in Figure~\ref{fig:newLabels}. We then check its performance. See Figure~\ref{fig:newLabelsMetrics}. As you can see, the classification metric does not change in most of cases. This means our method is robust to keywords change. This is because we leverage semantic information using word embedding trained on the dataset. And negative sampling helps our model identify words semantically related keywords. This helps our method leverage more information beyond bag of word representations. This experiment shows that our method can perform well when the input keywords is less common in corpus.
This section continues the ablation studies reportede in the main paper.
\subsection{Effect of Increase in temperature} \label{effect_distribution_temp}
\begin{figure*}
\hspace*{-9em}\includegraphics[scale=0.55]{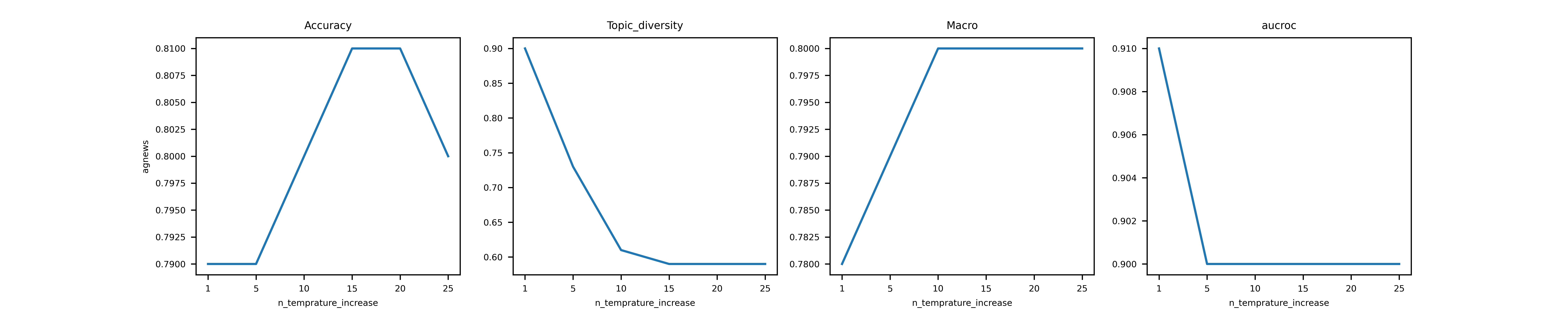}
\caption{Impact of increasing temperature of vMF VS various metrics on AG News for S2vNTM model.}
\label{fig:ntemprature_exp}
\end{figure*}

Temperature is the constant multiplied to the sampled distribution from vMF before softmax. Because of this trick, the topic distribution become more representative and therefore it becomes easier for the model to identify those topics or clusters. 
Now if the temperature is too high, the distance between topic clusters will increase and the model will have difficulty in adjusting clusters based on keywords set since keywords may be far from each other in latent space. This can be observed in Figure~\ref{fig:ntemprature_exp} when the temperature is increased from 20 and above we see a decrease in accuracy. On the contrary, if the temperature is too small, the latent distribution is less representative which makes the boundary between clusters vague. This again decreases the performance of the model. This can be observed in Figure~\ref{fig:ntemprature_exp} from values 5 to 15. The benefits of lower temperature is to make topic more diverse as you can see in second Figure~\ref{fig:ntemprature_exp}. The optimal value for temperature given number of topics (=5 for this experiment) and the dataset is anywhere between 15-20 where model can easily identify topics. The temperature within this range has high clusterability and expressibility.

\subsection{Effect of Gamma} 
\begin{figure*}
\hspace*{-11em}\includegraphics[scale=0.57]{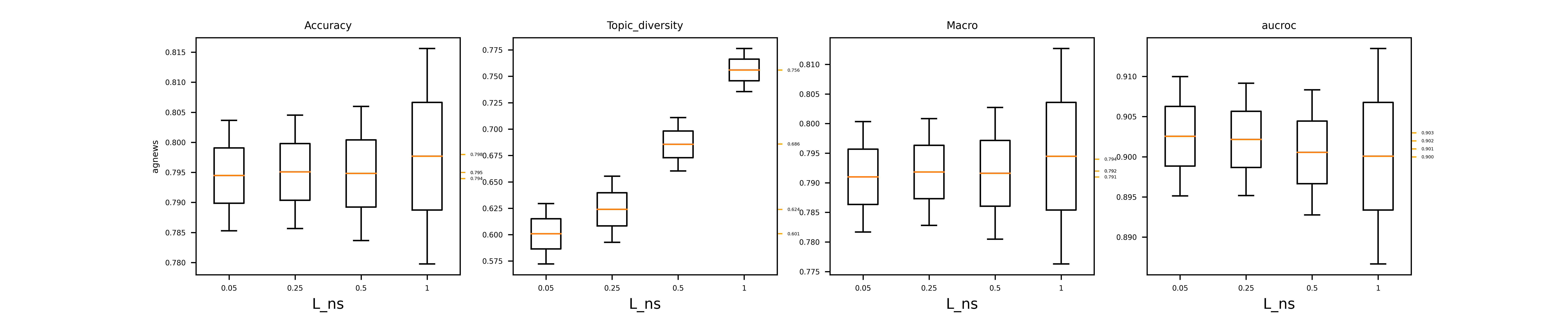}
\caption{Effect of gamma. (y-axis on right shows mean.)}
\label{fig:gamma_effect}
\end{figure*}

We explore the effect of various values of gamma shown in Figure~\ref{fig:gamma_effect}. With the increase in gamma, we observe a minimal increase in standard deviation and mean in accuracy and macro. Higher gamma makes $L_{NS}$ stronger which makes the model less stable.  We observe a strong increase in diversity score. This is because higher gamma score can push unrelated keywords further away. This makes each topic more coherent and different from other topics. So, at higher gamma, there is significant increase in diversity with negligible sacrifice in accuracy. This indicates stability of the method.

\subsection{Effect of batch normalization and dropout}
\label{dropout}
\begin{figure*}
\hspace*{-14em}\includegraphics[scale=0.45]{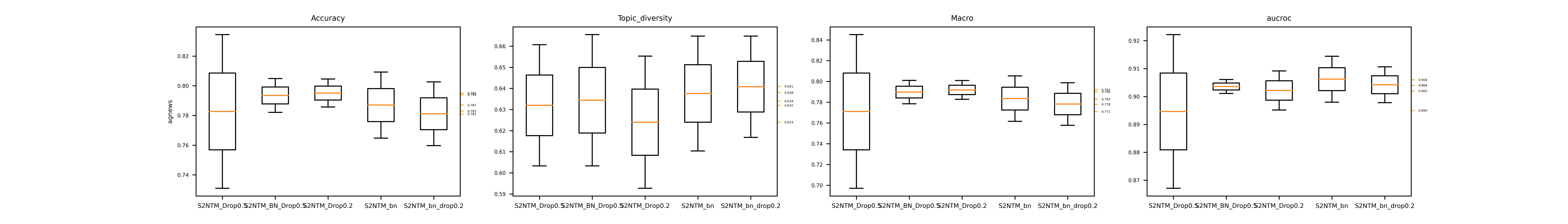}

\caption{Effect of Batch Normalization and Dropout. (y-axis on right shows mean.)}
\label{fig:BN_drop_effect}
\end{figure*}

We explore various combinations of batch normalization (bn) and dropout which are shown in Figure~\ref{fig:BN_drop_effect}. Independently, S2vNTM{\_}Drop0.5 i.e. S2vNTM with dropout 0.5 has high standard deviation. Reducing dropout to 0.2 S2vNTM{\_}Drop0.2 or adding bn S2vNTM{\_}bn{\_}Drop0.5 have very similar effect of reduced variance for accuracy and Macro F1  but S2vNTM{\_}bn{\_}Drop0.5 has higher aucroc and diversity and less variance. In general, adding bn with dropout stabilizes the model performance which was expected.

\subsection{Effect of learnable distribution temperature}
In Appendix~\ref{effect_distribution_temp} we discuss effect of increasing distribution temperature. In this study, we make it a learnable parameter and implement it in two ways. The first way is setting temperature variable as one parameter that can be learned (1-p model). All topics share the same parameter. The second way is setting temperature variable as a vector with dimension equal to the number of topics (n-p model). This means each topic has its own temperature. The initialization value for both the vectors is 10.

After training, the 1-p model has value 4.99 and n-p model has values [-0.45,4.88,5.91,3.47,4.19] (values are rounded to 2 decimals). The accuracy for 1-p model is 78.9 and n-p model is 80.5. This means that our method can further improve with learnable temperature.

In Appendix~\ref{effect_distribution_temp} we found that distribution temperature values between 15 to 20 gave highest accuracy (81) but on the contrary the learned values in 1-p is 4.99 (accuracy 78.9). This means that our loss function is not fully aligned with accuracy metric. This is due to the fact that we optimize reconstruction loss as well as KL divergence during the training procedure. This makes our objective less aligned with cross entropy loss.

\clearpage

\begin{figure*}
\hspace*{-7em}\includegraphics[trim=10 20 30 40, clip, scale=0.62]{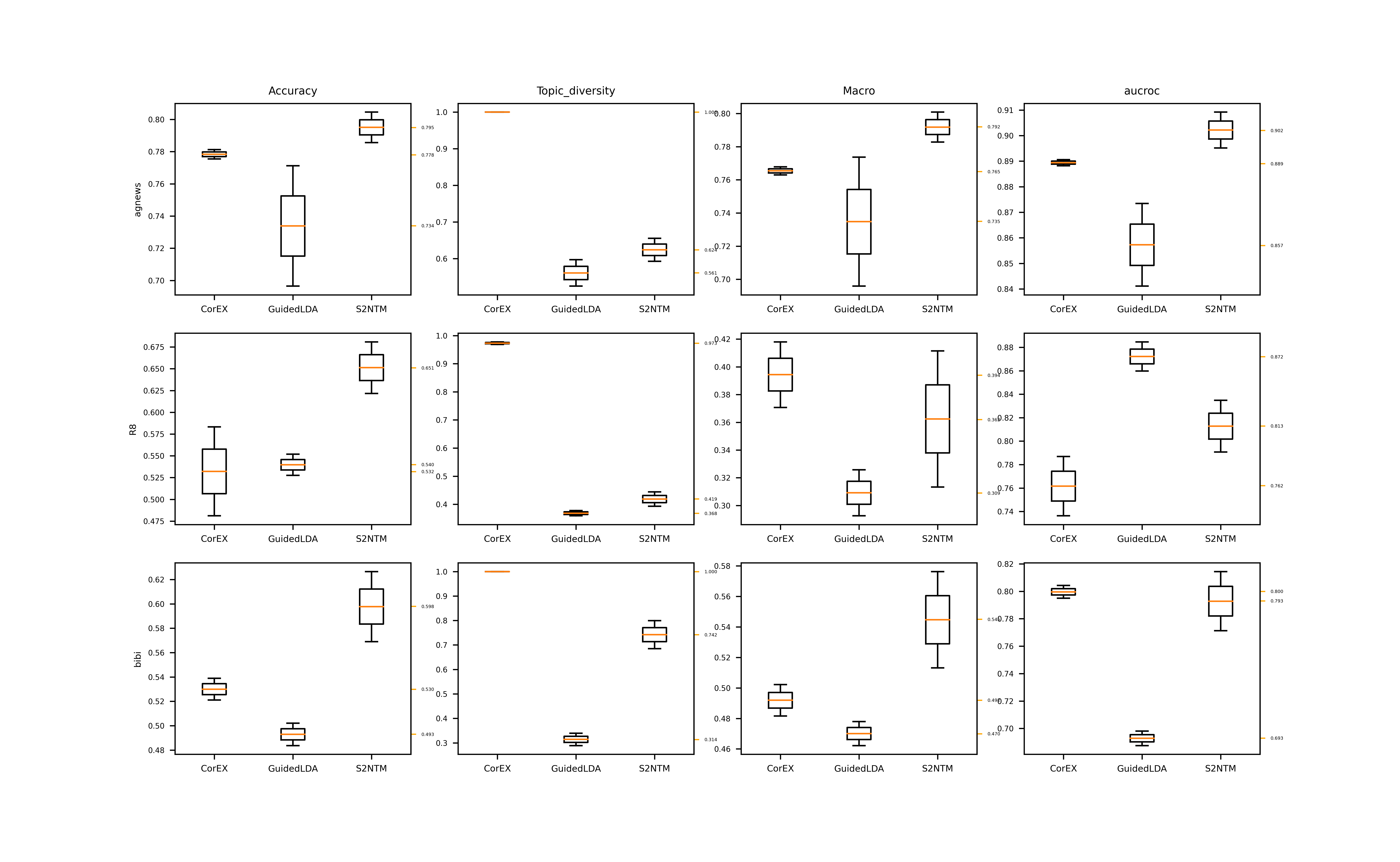}
\caption{Results for Accuracy, Topic Diversity, Macro F1 and aucroc for GuidedLDA, CoreEx and S2vNTM. (Right y-axis shows mean).}
\label{fig:main_metrics}
\end{figure*}

\begin{table*}[h!]
\centering
\begin{tabular}{ |c|c|c|c| } 
\hline
GuidedLDA & CorEx \\
\hline
iraq, kill, reuters, president, minister &
\textbf{government}, \textbf{war}, \textbf{military}, iraq, kill \\
\hline
reuters, \textbf{stock}, oil, price, profit &
\textbf{stock}, \textbf{market}, \textbf{industry}, price, oil \\
\hline
microsoft, company, \textbf{software}, service, internet & \textbf{software}, \textbf{computer}, microsoft, internet, service \\
\hline
win, game, team, season, lead & \textbf{footable}, \textbf{basketball}, game, win, season
\\
\hline
space, reuters, win, quot, world &
court, executive, chief, commission, union\\
\hline
quot, year, company, million, plan & 
inc, corp, people, bush, million\\
\hline
\end{tabular}

\caption{Compare top 5 keywords from each topics for GuidedLDA and CorEx using Dataset AG News. The keywords that are given is [government,military,war], [stock,market,industry], [computer,telescope,software], [basketball,football,athlete]. CorEx provided diverse keywords but they are not similar in meaning which can make users confused.}
\label{tb_loss:table_of_figures2}
\end{table*}

\end{document}